\documentclass[journal]{IEEEtran}

\usepackage{graphicx} 
\usepackage{supertabular}
\usepackage{longtable}
\usepackage{subfloat}

\ifCLASSINFOpdf
\else
\fi
\usepackage{amssymb}
\usepackage{amsthm}
\usepackage{amsmath}
\usepackage{algorithmic}
\usepackage{algorithm}
 \usepackage{tabularx}

\begin{document}
%
\title{On Extreme Pruning of Random Forest Ensembles for Real-time Predictive Applications}
\author{Khaled Fawagreh,
        Mohamed Medhat Gaber,
        and~Eyad~Elyan,~\IEEEmembership{Member,~IEEE}
\thanks{K. Fawagreh, M. M. Gaber, and E. Elyan are with
IDEAS Research Institute, Robert Gordon University, Riverside East, Garthdee Road, Aberdeen
AB10 7GJ, UK (k.fawagreh@rgu.ac.uk, m.gaber1@rgu.ac.uk, e.elyan@rgu.ac.uk)}}
{Shell \MakeLowercase{\textit{et al.}}: Bare Demo of IEEEtran.cls for Journals}
%



\maketitle

\begin{abstract}
Random Forest (RF) is an ensemble supervised machine learning technique that was developed by Breiman over a decade ago. Compared with other ensemble techniques, it has proved its accuracy and superiority. Many researchers, however, believe that there is still room for enhancing and improving its performance accuracy. This explains why, over the past decade, there have been many extensions of RF where each extension employed a variety of techniques and strategies to improve certain aspect(s) of RF. Since it has been proven empirically that ensembles tend to yield better results when there is a significant diversity among the constituent models, the objective of this paper is twofold. First, it investigates how data clustering (a well known diversity technique) can be applied to identify groups of similar decision trees in an RF in order to eliminate redundant trees by selecting a representative from each group (cluster).  Second, these likely diverse representatives are then used to produce an extension of RF termed \emph{CLUB-DRF} that is much smaller in size than RF, and yet performs at least as good as RF, and mostly exhibits higher performance in terms of accuracy. The latter refers to a known technique called ensemble pruning. Experimental results on 15 real datasets from the UCI repository prove the superiority of our proposed extension over the traditional RF. Most of our experiments achieved at least 95\% or above pruning level while retaining or outperforming the RF accuracy. 

\end{abstract}


%
\IEEEpeerreviewmaketitle

\section{Introduction}
\label{intro}

\IEEEPARstart{E}nsemble classification is an application of ensemble learning to boost the accuracy of classification. Ensemble learning is a supervised machine learning paradigm where multiple models are used to solve the same problem \cite{polikar2006ensemble} \cite{rokach2010ensemble} \cite{kuncheva2003measures}. Since single classifier systems have limited predictive performance \cite{yan2004designing} \cite{polikar2006ensemble} \cite{maclin2011popular}  \cite{rokach2010ensemble}, ensemble classification was developed to yield better predictive performance \cite{polikar2006ensemble} \cite{maclin2011popular}  \cite{rokach2010ensemble}. In such an ensemble, multiple classifiers are used. In its basic mechanism, majority voting is then used to determine the class label for unlabeled instances where each classifier in the ensemble is asked to predict the class label of the instance being considered. Once all the classifiers have been queried, the class that receives the greatest number of votes is returned as the final decision of the ensemble.

Three widely used ensemble approaches could be identified, namely, boosting, bagging, and stacking. Boosting is an incremental process of building a sequence of classifiers, where each classifier works on the incorrectly classified instances of the previous one in the sequence. AdaBoost \cite{freund1997decision} is the representative of this class of techniques. However, AdaBoost is prone to overfitting. The other class of ensemble approaches is the Bootstrap Aggregating (Bagging) \cite{breiman1996baggingrandom}. Bagging involves building each classifier in the ensemble using a randomly drawn sample of the data, having each classifier giving an equal vote when labeling unlabeled instances. Bagging is known to be more robust than boosting against model overfitting. Random Forest (RF) is the main representative of bagging  \cite{breiman2001random}. Stacking (sometimes called stacked generalization) extends the cross-validation technique that partitions the dataset into a held-in dataset and a held-out dataset; training the models on the held-in data; and then choosing whichever of those trained models performs best on the held-out data. Instead of choosing among the models, stacking combines them, thereby typically getting performance better than any single one of the trained models \cite{wolpert1992stacked}. Stacking has been successfully used on both supervised learning tasks (regression) \cite{breiman1996stacked}, and unsupervised learning (density estimation) \cite{smyth1999linearly}. 

The ensemble method that is relevant to our work in this paper is RF. RF has been proved to be the state-of-the-art ensemble classification technique. Since RF algorithms typically build between 100 and 500 trees \cite{williams2011use}, in real-time applications, it is of paramount importance to reduce the number of trees participating in majority voting and yet achieve performance that is at least as good as the original ensemble. In this paper, we propose a data clustering approach to prune RF ensembles where only a small subset of the ensemble is selected. We cluster trees according to their classification behavior on a subset of the dataset. Then we choose only one tree from each cluster, motivated by the fact that the tree is a representative of its cluster. At voting time, the number of voting trees is reduced significantly yielding classification accuracy at least as good as all voting trees. A thorough experimental study is conducted, with multiple of 5 clusters ranging from 5 through to 40, over an RF of 500 trees. Results show the potential of our technique for deployment in real-time systems, yielding higher predictive accuracy than traditional RF, with 17 to 100 times faster classification per instance resulting from pruning levels in the range of 94\%--99\%. 

This paper is organized as follows. First we discuss related work in Section \ref{related}. This is followed by Section \ref{prem} where preliminaries about the motivation and introduction to RF are covered.  In Section \ref{cluster}, we present the clustering technique that will be utilized in our approach. Section \ref{extend} formalizes our approach and corresponding algorithm. Experiments, results and analysis demonstrating the superiority of the proposed pruned RF over the traditional RF are detailed in Section \ref{experiment}. The paper is finally concluded with a summary and pointers to future directions in Section \ref{conc}. 

\section{Related Work}
\label{related}

Several enhancements have been made in recent years in order to produce a subset of an ensemble  that performs as well as, or better than, the original ensemble. The purpose of ensemble pruning is to search for such a good subset. This is particularly useful for large  ensembles that require extra memory usage, computational costs, and occasional decreases in effectiveness.  Grigorios et al. \cite{tsoumakas2009ensemble} recently amalgamated a survey of ensemble pruning techniques where they classified such techniques into four categories: ranking based, clustering based, optimization based, and others. Clustering based methods, that are relevant to us in  this paper,  consist of  two stages. In the first stage, a clustering algorithm is employed in order to discover groups of models that make similar predictions. Pruning each cluster then follows in the final stage. In this stage, several approaches have been used.  One approach by \cite{bakker2003clustering} was to train a new model for each cluster, using the cluster centroids as values of the target variable. Another interesting approach was proposed by \cite{giacinto2000design} that involved selecting from each cluster the classifier that is most distant to the rest of the clusters. A yet different approach by \cite{lazarevic2001effective} that does not guarantee the selection of a single model from each cluster was by iteratively removing models from the least to the most accurate, until the accuracy of the entire ensemble starts to decrease. Selecting the most accurate model from each cluster was proposed by \cite{qiang2005clustering}. 

It is worth pointing out that there are several techniques available in the literature that use clustering-based approaches to reduce the number of trees in the ensemble (see \cite{EE6282329} for a good review). Unlike such techniques, our clustering approach in this paper is different in many aspects. First of all, to the best of our knowledge, none of the clustering-based approaches was developed for RF ensembles. For example, all the approaches proposed by the respective anthers in \cite{bakker2003clustering}  \cite{giacinto2000design} \cite{lazarevic2001effective} \cite{qiang2005clustering} were developed for neural network ensembles. Furthermore, our approach differs from theirs in two other ways. For the selection of a representative from each cluster, we have selected the best performing representative on the out-of-bag (OOB) instances and this selection method was not used by anyone of them. As will be discussed in Section \ref{extend}, using OOB samples to evaluate a tree gives an unbiased estimate of its predictive accuracy since, unlike about 64\% of the training data that was seen by the tree when it was built, OOB data was not seen and therefore, it is a more accurate measure of the tree's  predictive accuracy. Furthermore, at the experimental level, we have used 15 datasets from the UCI repository, however, very few datasets were used by them:  2 in \cite{bakker2003clustering}, 1 in  \cite{giacinto2000design}, 4 in \cite{lazarevic2001effective}, and 4 in \cite{qiang2005clustering}.

Recent work in ensemble pruning  just this year (2014) alone was reported.  Without a significant loss of prediction accuracy, a combination of static and dynamic pruning techniques were applied on Adaboost ensembles in order to yield less memory consumption and improved classification speed \cite{soto2014double}. A pruning scheme for high dimensional and large sized benchmark datasets was developed by \cite{diao2014feature}, In such a scheme,  an extended feature selection technique was used to transform ensemble predictions into training samples, where classifiers were treated as  features. Then a global heuristic harmony search was used to select a smaller subset of such artificial features.

\subsection{Diversity Creation Methods}
Because of the vital role diversity plays on the performance of ensembles, it had received a lot of attention from the research community. G. Brown et al. \cite{brown2005diversity} summarized the work done to date in this domain from two main perspectives. The first is a review of the various attempts that were made to provide a formal foundation of diversity. The second, which is more relevant to this paper, is a survey of the various techniques to produce diverse ensembles. For the latter, two types of diversity methods were identified: implicit and explicit. While implicit methods tend to use randomness to generate diverse trajectories in the hypothesis space, explicit methods, on the other hand, choose different paths in the space deterministically. In light of these definitions, bagging and boosting in the previous section are classified as implicit and explicit respectively. 

G. Brown et al. \cite{brown2005diversity} also categorized ensemble diversity techniques into three categories: starting point in hypothesis space, set of accessible hypotheses, and manipulation of training data. Methods in the first category use different starting points in the hypothesis space, therefore, influencing the convergence place within the space.  Because of their poor performance of achieving diversity, such methods are used by many authors as a default benchmark for their own methods \cite{maclin2011popular}. Methods in the second category vary the set of hypotheses that are available and accessible by the ensemble. For different ensembles, these methods vary either the training data used or the architecture employed. In the third category, the methods alter the way space is traversed. Occupying any point in the search space, gives a particular hypothesis. The type of the ensemble obtained will be determined by how the space of the possible hypotheses is traversed.

\subsection{Diversity Measures}
Regardless of the diversity creation technique used, diversity measures were developed to measure the diversity of a certain technique or perhaps to compare the diversity of two techniques. Tang et al. \cite{tang2006analysis} presented a theoretical analysis on six existing diversity measures: disagreement measure  \cite{skalak1996sources}, double fault measure  \cite{giacinto2001design}, KW variance \cite{kohavi1996bias}, inter-rater agreement \cite{fleiss2013statistical}, generalized diversity \cite{partridge1997software}, and measure of difficulty \cite{fleiss2013statistical}. The goal was not only to show the underlying relationships between them, but also to relate them to the concept of margin, which is one of the contributing factors to the success of ensemble learning algorithms.

We suffice to describe the first two measures as the others are outside the scope of this paper. The disagreement measure is used to measure the diversity between two base classifiers (in RF case, these are decision trees) \emph{$h_j$} and \emph{$h_k$}, and is calculated as follows:

$$
dis_{j,k} = \frac{N^{10}+N^{01}}{N^{11}+N^{10}+N^{01}+N^{00}}
$$
\\
where \\
\begin{itemize}
\item \emph{$N^{10}$}: means number of training instances that were correctly classified by \emph{$h_j$}, but are incorrectly classified by \emph{$h_k$}
\item \emph{$N^{01}$}: means number of training instances that were incorrectly classified by \emph{$h_j$}, but are correctly classified by \emph{$h_k$} 
\item \emph{$N^{11}$}: means number of training instances that were correctly classified by  \emph{$h_j$} and  \emph{$h_k$}
\item \emph{$N^{00}$}: means number of training instances that were incorrectly classified by  \emph{$h_j$} and  \emph{$h_k$} 
\end{itemize}
The higher the disagreement measure, the more diverse the classifiers are. The double fault measure uses a slightly different approach where the diversity between two classifiers is calculated as:
$$
DF_{j,k} = \frac{N^{00}}{N^{11}+N^{10}+N^{01}+N^{00}}
$$
 The above two diversity measures work only for binary classification (AKA binomial) where there are only two possible values (like Yes/No) for the class label, hence, the objects are classified into exactly two groups.  They do not work for multiclass (AKA multinomial) classification where the objects are classified into more than two groups. In Section \ref{dmeasure}, we propose a simple diversity measure that works with both binary and multiclass classification.

\section{Preliminaries}
\label{prem}
\subsection{Motivation}
\label{motive}
As mentioned before, RF algorithms tend to build between 100 and 500 trees \cite{williams2011use}. Some empirical and theoretical studies have also clearly demonstrated that adding more trees to an RF beyond a certain number (i.e. 500) won't necessarily improve the RF accuracy \cite{EE5178693}. Our research aims at pruning RF ensembles by  producing a subset of the original ones that are significantly smaller in size and yet, have accuracy performance that is at least as good as that of the original RF from which they were derived. In other words, we are aiming at finding the optimial or near-optimal number of trees that will be used to generate an accurate RF.

\subsection{Random Forest}
\label{rf}
Random Forest is an ensemble supervised machine learning technique used for classification and regression. Developed by Breiman \cite{breiman2001random}, the method combines Breiman's bagging sampling approach \cite{breiman1996baggingrandom}, and the random selection of features, introduced independently by Ho \cite{ho1995random} \cite{ho1998random} and Amit and Geman \cite{amit1997shape}, in order to construct a collection of decision trees with controlled variation. Using bagging,  each decision tree in the ensemble is constructed using a sample with replacement from the training data. Statistically, the sample is likely to have about 64\% of instances appearing at least once in the sample. Instances in the sample are referred to as in-bag-instances, and the remaining instances (about 36\%), are referred to as out-of-bag instances. Each tree in the ensemble acts as a base classifier to determine the class label of an unlabeled instance. This is done via majority voting where each classifier casts one vote for its predicted class label, then the class label with the most votes is used to classify the instance. Algorithm \ref{rfalgo} below depicts the RF algorithm \cite{breiman2001random} where N is the number of training samples and S is the number of features in dataset.

\begin{algorithm}[!htb]
\caption{Random Forest Algorithm}          
\label{rfalgo}                           
\begin{algorithmic}
          
\STATE \COMMENT{User Settings}
\STATE input $N$, $S$   
\STATE \COMMENT{Process}
\STATE Create an empty vector $\overrightarrow{RF}$ 
\FOR{$i = 1 \to N$}
\STATE Create an empty tree $T_i$
\REPEAT
\STATE Sample $S$  out of all features $F$ using Bootstrap sampling 
\STATE Create a vector of the $S$ features $\overrightarrow{F_S}$
\STATE Find Best Split Feature $B(\overrightarrow{F_S})$
\STATE Create A New Node using $B(\overrightarrow{F_S})$ in $T_i$
\UNTIL{No More Instances To Split On}
\STATE Add $T_i$ to the $\overrightarrow{RF}$ 
\ENDFOR
\STATE \COMMENT{Output}
\STATE A vector of trees $\overrightarrow{RF}$
\end{algorithmic}
\end{algorithm}

To achieve optimal results, the classifiers in the ensemble should both be accurate and diverse. An accurate classifier is one that has an error rate better than random guessing.  Two classifiers are diverse if they make different errors on new data points. The more diverse the classifiers are, the better the results are. In fact, it has been proven empirically that ensembles tend to yield better results when there is a significant diversity among the models \cite{kuncheva2003measures} \cite{brown2005diversity} \cite{adeva2005accuracy}  \cite{tang2006analysis}. This explains why many ensemble methods seek to promote diversity among the models they combine. In this paper, we will employ a data clustering technique to produce ensembles with diverse trees.

\section{Clustering}
\label{cluster}
Clustering has been used extensively as a diversity technique in many applications \cite{li2010clustering} \cite{kuncheva2004using} \cite{brown1998evaluation} \cite{shemetulskis1995enhancing}  \cite{lee2008cluster} \cite{sharpton2012sifting}. Unlike classification, clustering is an unsupervised learning technique that attempts to organize objects into groups whose members are similar in some way. Each group is referred to as a cluster, hence, a cluster is a collection of objects which are similar between them and are dissimilar to the objects belonging to other clusters. Clustering is considered a data exploration method as it helps to unveil the natural grouping in a dataset without a prior knowledge of the groups to be produced. One of the earliest and most popular clustering algorithms is called K-means. It was developed by MacQueen \cite{macqueen1967some} in the late sixties and despite its seniority, it is still considered as one of the most widely used algorithms, mainly due to its simplicity, efficiency, and empirical success \cite{jain2010data}.

Unfortunately, however, this algorithm has  a limitation that it only works with numerical data.To overcome this limitation, there have been some extensions of this algorithm to work with categorical data \cite{huang1998extensions} \cite{huang1999fuzzy} \cite{san2004alternative}. Huang \cite{huang1998extensions} developed an extension of K-means called K-modes that uses modes instead of means, and can handle categorical data using the following simple matching dissimilarity measure:

\begin{equation}
      d_{1}={\sum\limits^{m}_{j=1}}\delta(x_{j},y_{j})
\end{equation}  
where X, Y be two categorical objects with \emph{m} categorical attributes, $x_{j}$ and $y_{j}$ (j=1..m) refer to the categorical attributes of X and Y respectively and

\begin{equation}
    \delta(x_{j},y_{j})=
    \begin{cases}
      0, & \text{if}\ x_{j} = y_{j} \\
      1, & \text{otherwise}
    \end{cases}
  \end{equation}

According to the above dissimilarity measure, the total number of mismatches of the corresponding attribute categories of the two objects is calculated. The similarity of the two objects is conversely proportional to the number of mismatches; the smaller it is, the more similar the two objects are. Algorithm \ref{kmodes} outlines the main steps involved in the K-modes algorithm.

\begin{algorithm}[!htb]
\caption{K-modes Algorithm}          
\label{kmodes}                           
\begin{enumerate}
\item Select k initial modes, one for each cluster.
\item Allocate an object to the cluster whose mode is the nearest to it according to (1). Update
the mode of the cluster after each allocation.
\item After all objects have been allocated to clusters, retest the dissimilarity of objects against
the current modes. If an object is found such that its nearest mode belongs to another
cluster rather than its current one, reallocate the object to that cluster and update the
modes of both clusters.
\item Repeat 3 until no object has changed clusters after a full cycle test of the whole data set.
\end{enumerate}
\end{algorithm}

In the experimental stage we will be using a popular machine learning software suite called Waikato Environment for Knowledge Analysis (WEKA) \cite{witten2005data}. This suite comes with a clustering algorithm that integrates K-means and K-modes to work with both numerical and categorical data.   

\section{Proposed Extension}
\label{extend}
In this section, we propose an extension of RF called CLUB-DRF that spawns a child RF that is 1) much smaller in size than the parent RF and 2) has an accuracy that is at least as good as that of the parent RF. In the remainder of this paper, we will refer to the parent/original traditional Random Forest as \emph{RF}, and refer to the resulting child RF based on our method as \emph{CLUB-DRF}. 

\subsection{CLUstering-Based Diverse Random Forest (CLUB-DRF)}
\label{CLUB-DRF}

\begin{figure}[t]
\centering
 \includegraphics[scale=.3]{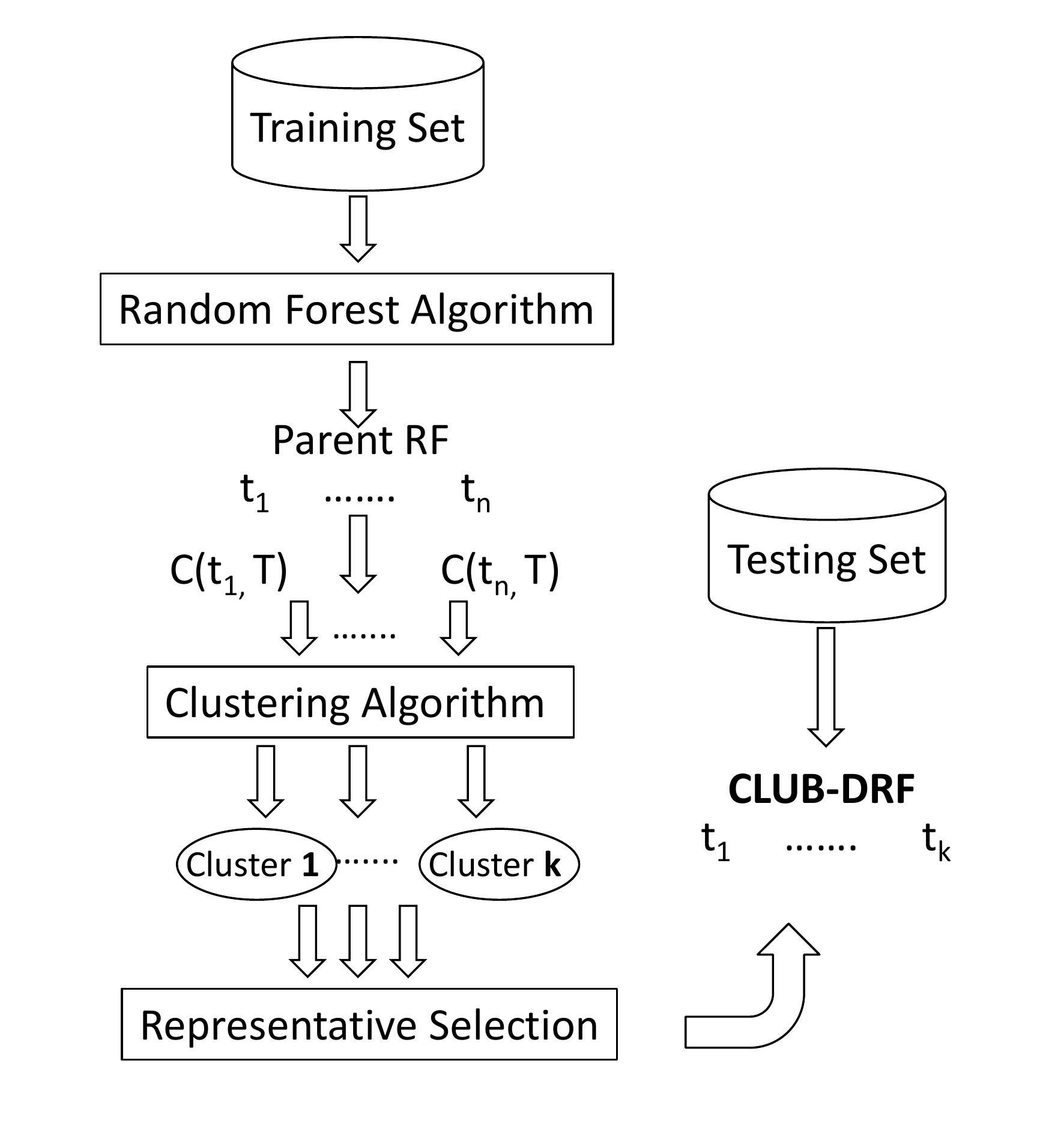}
\caption{CLUB-DRF Approach}
\label{clubdrf}
\end{figure}

The CLUB-DRF extension applies a clustering-based technique to produce diverse groups of trees in the RF.   Assuming that the trees in the RF are denoted by the vector RF=$<t_{1}$, $ t_{2}$,$\cdots$,$t_{n}>$ (where n is number of trees in the RF), and the training set is denoted by T=\{$r_1$,$ r_2$,$\cdots$,$r_m$\}. Each tree in the RF will then be used to classify each record in the training set to determine the class label c. We use C( $t_i$, T) (where $t_i$ $\in$ RF) to denote a vector of class labels obtained  after having  $t_i$ classify the training set T. That is  \\  C( $t_i$, T)=$<c_{i1}$, $ c_{i2}$,$\cdots$,$c_{im}>$. The result obtained of having each tree classify the training records will therefore be a super vector $\AA$ containing class labels vectors as classified by each tree (that is, $\AA$ is a vector of vectors):

$\AA$  = C( $t_1$, T) $\cup$ C( $t_2$, T) $\cup$ ... C( $t_n$, T) 
\\ This set will be fed as input to a clustering algorithm as shown in Figure \ref{clubdrf}. When clustering is completed, we will have a set of clusters where each cluster contains vectors that are similar and likely to have the least number of discrepancies. For example, using a training set of 5 records with the class label being a boolean (Y/N),  the vectors $<Y,Y,Y,N,N>$ and $<Y,Y,Y,Y,N>$ are likely to fall in the same cluster. However, the vectors $<Y,Y,Y,N,N>$ and $<Y,N,N,Y,Y>$ are likely to appear in different clusters because there are many discrepancies in the class labels. As defined above, since a cluster contains objects which are similar between them but are dissimilar to other objects belonging to other clusters, a vector of class labels as classified by a tree in one cluster will be dissimilar (different) from another vector belonging to another cluster, and this means that the two vectors are diverse.
 
When using a clustering algorithm that requires the number of clusters to be specified in advance (as in K-means), one interesting and challenging question that immediately comes to mind is how to determine this number so that it is not high and it is not low.  Mardi \cite{mardia1980multivariate} proposed a simple rule of thumb:
\emph{number\_of\_clusters}  $\approx$ $\sqrt{\frac{n}{2}}$, where \emph{n} refers to the size of data points to be clustered.

Based on Figure \ref{clubdrf}, we formalize the CLUB-DRF algorithm as shown in Algorithm \ref{club-drf-algo} where T stands for the training set. The constant k refers to the number of clusters to be created which we define as a multiple of 5  in the range 5 to 40. This way and as we shall see in the experiments section, we can compare the performance of an CLUB-DRF of different sizes with that of the RF.

It is important to remember that the size of the resulting CLUB-DRF is determined by the number of clusters used. For example,  if the number of the clusters is 5, then the resulting CLUB-DRF will have size 5, and so on. 

\begin{algorithm}[!htb]
\caption{CLUB-DRF Algorithm}          
\label{club-drf-algo}                           
\begin{algorithmic}
          
\STATE \COMMENT{User Settings}
\STATE input $T$, $k$
\STATE \COMMENT{Process}
\STATE Create an empty vector $\overrightarrow{classLabels}$ 
\STATE Create an empty vector $\overrightarrow{CLUB-DRF}$ 
\STATE Using \emph{T}, call Algorithm $\ref{rfalgo}$ above to create  \emph{RF} 
\FOR{$i = 1 \to RF.getNumTrees()$}
\STATE  $\overrightarrow{classLabels}$ $\Leftarrow$  $\overrightarrow{classLabels}$ $\cup$ C(RF.tree(i), T)
\ENDFOR
\STATE Cluster $\overrightarrow{classLabels}$ into a set of \emph{k} clusters: \\ $cluster_1$ ... $cluster_k$  
\STATE From each cluster, select a representative tree RF.tree(j) that corresponds to the instance C(RF.tree(j), T) in the cluster \\
\STATE Add $RF.tree(j)$ to $\overrightarrow{CLUB-DRF}$ 
\STATE \COMMENT{Output}
\STATE A vector of trees $\overrightarrow{CLUB-DRF}$
\end{algorithmic}
\end{algorithm}

When selecting a representative from each cluster (refer to Algorithm \ref{club-drf-algo} above), we will consider the best performing representative  on the out-of-bag (OOB) instances. As mentioned in Section \ref{rf}, these are the instances that were not included in the sample with replacement that was used to build the tree, and they account for about 36\% of the total number of instances.  Using the OOB samples to evaluate a tree gives an unbiased estimate of its predictive accuracy since, unlike training data that was seen by the tree when it was built, OOB data was not seen and therefore, it is a more accurate measure of the tree's  predictive accuracy. 

\subsection{Diversity Measure}
\label{dmeasure}

Here we propose a simple diversity measure to measure the diversity of two classifiers that works with binary and multiclass classification.  Given two classifiers \emph{$h_j$} and \emph{$h_k$} and a training set T of size \emph{n}. Let $C$($t_l$,$s_i$)  denote the class label obtained  after having  $t_l$ classify the sample $s_i$ in the training set T. Utilizing the dissimilarity measure defined in (1) above, the diversity between the two classifiers can be measured by: 
\begin{equation}
diversity_{j,k}=\frac{\sum\limits^{n}_{i=1}\delta(C( t_j,v_i) ,C( t_k,v_i) )}{n} 
\end{equation}
The higher the number of discrepancies between the two classifiers, the higher the diversity is. For example, assume that we have a training set consisting of 10 training samples
T=\{$s_1$,$ s_2$,$ s_3$,$ s_4$,$ s_5$,$ s_6$,$ s_7$,$ s_8$,$ s_9$,$s_{10}$\}, and two classifiers $t_1$ and $t_2$. Assume also that there are 3 possible values for the class label \{a,b,c\}. Let  C($t_1$,T)=\textless a,a,b,c,c,a,b,c,b,b\textgreater  
\hphantom aand C($t_2$,T)=\textless a,a,b,b,a,a,b,c,c,c\textgreater. According to (3) above, the diversity between the two classifiers is therefore 4/10 or 40\%.

\section{Experimental Study}
\label{experiment}

For our experiments, we have used 15 real datasets with varying characteristics from the UCI repository \cite{Bache+Lichman:2013}. To use the holdout testing method, each dataset was divided into 2 sets: training and testing. Two thirds (66\%) were reserved for training and the rest (34\%) for testing.  Each dataset consists of input variables (features) and an output variable. The latter refers to the class label whose value will be predicted in each experiment.  For the RF in Figure \ref{clubdrf}, the initial RF to produce the CLUB-DRF had a size of 500 trees, a typical upper limit setting for RF \cite{williams2011use}. We chose the upper limit for two main reasons. First, the more trees we have, the more diverse ones we can get. Secondly, when we have many clusters, the more trees we have, the more unlikely that we wind up with empty clusters \cite{pakhira2009modified}. 

The CLUB-DRF algorithm described above was implemented using the Java programming language utilizing the API of Waikato Environment for Knowledge Analysis (WEKA) \cite{witten2005data}. We ran the CLUB-DRF algorithm 10 times on each dataset where a new RF was created in each run. We then calculated the average of the 10 runs for each resulting CLUB-DRF to produce the average for a variety of metrics including accuracy rate, minimum accuracy rate, maximum accuracy rate, standard deviation, FMeasure, and AUC as shown in Table \ref{bestrepoobtable}.  For the RF, we just calculated the average accuracy rate, FMeasure, and AUC as shown in the last 3 columns of the table.

\subsection{Results} 
\label{result}

Table \ref{bestrepoobtable} compares the performance of CLUB-DRF and RF on the 15 datasets. To show the superiority of CLUB-DRF, we have highlighted in boldface the average accuracy rate of CLUB-DRF when it is greater than that of RF, and underlined the accuracy rate when it is equal to that of RF (only one occurrence found in the \emph{squash-unstored} dataset).
Taking a closer look at this table, we find that CLUB-DRF performed at least as good as RF on 13 datasets. Interestingly enough, of the 13 datasets, CLUB-DRF, regardless of its size, completely outperformed RF on 6 of the datasets, namely, \emph{breast-cancer}, \emph{pasture},  \emph{eucalyptus},  \emph{glass}, \emph{sonar}, and \emph{vehicle}. 
While CLUB-DRF lost to RF on only 2 datasets (\emph{audit} and  \emph{vote}), the difference was by a very small negligible fraction of less than 1\%!

\subsection{Analysis} 
\label{analysis}
 
\begin{figure}[H]
\scalebox{0.60}{
\centering
\includegraphics[width=6in,height=4in]{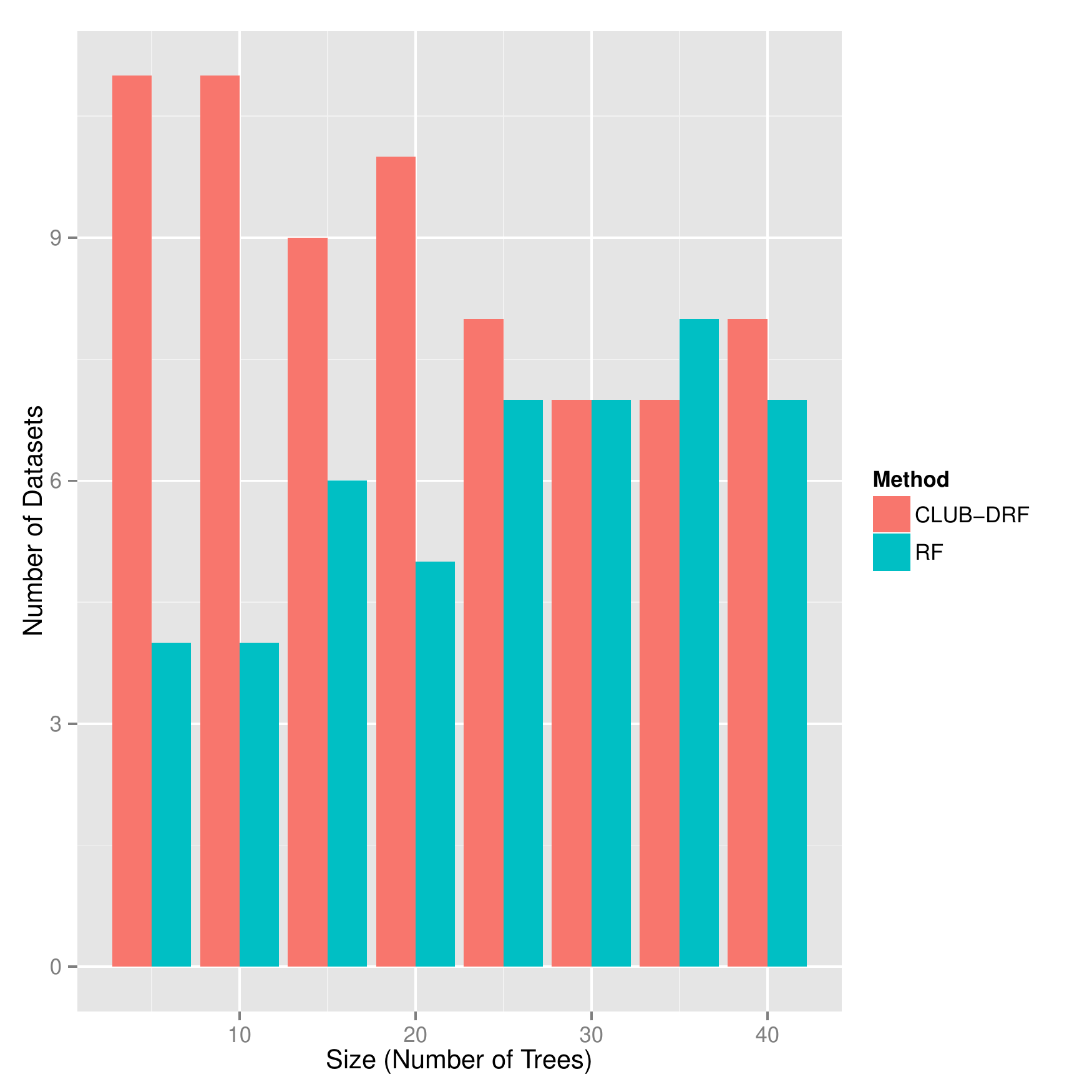}}
\caption{Accuracy Rate Comparison of RF \& CLUB-DRF}
\label{fig2}
\end{figure}

By showing the number of datasets each was superior on, Figure \ref{fig2} compares the accuracy rate of RF and CLUB-DRF using different sizes of CLUB-DRF. With one exception of size 35, the figure clearly shows that CLUB-DRF indeed performed at least as good as RF. Notably, the lesser the size of the ensemble, the better performance our method exhibits. This can be attributed to the higher diversity the ensemble has. Having 5 or 10 representatives guarantees a further away trees chosen to form the ensemble. When we move to a higher number of clusters, and consequently larger ensemble sizes, CLUB-DRF moves towards the original RF. This can be especially true when tree behaviors are no longer distinguishable, and creating more clusters does not add any diversity to the ensemble.


\subsection{Pruning Level}

By applying the above proposed clustering technique, we managed to achieve two objectives. First, CLUB-DRF ensembles with diverse trees were produced. Second and more importantly, we manged to significantly reduce the size of RF. The resulting pruned CLUB-DRF ensembles performed at least as good as the original RF, but mostly outperformed the original RF, as previously discussed. In ensemble pruning, a pruning level refers to the reduction ratio between the original ensemble and the pruned one. For example, if the size of the original ensemble is 500 trees and the pruned one is of size 50, then $100\% - \frac{50}{500} \times 100\%=90\%$ is the pruning level that was achieved in the pruned ensemble. This means that the pruned ensemble is 90\% smaller than the original one. Table \ref{oobPrunLevel} shows the pruning levels. 
The first column in  this table shows the maximum possible pruning level for an CLUB-DRF that has outperformed RF, and the second column shows the pruning level of the best performer CLUB-DRF. We can see that with extremely healthy pruning levels ranging from 94\% to 99\%, our technique outperformed RF. This makes CLUB-DRF a natural choice for real-time applications, where fast classification is an important desideratum. In most cases, 100 times faster classification can be achieved with the 99\% pruning level, as shown in the table. In the worst case scenario, only 16.67 times faster classification with 94\% pruning level in the \emph{white clover} dataset. Such estimates are based on the fact that the number of trees traversed in the RF is the dominant factor in the classification response time. This is especially true, given that RF trees are unpruned bushy trees.   


\begin{table}[ht]
\caption{Maximum Pruning Level with Best Possible Performance} 
\label{oobPrunLevel}
\scalebox{0.9}{
\centering 
\begin{tabular}{l c c c} 
\hline\hline 
Dataset & Maximum Pruning Level & Best Performance Pruning Level \\ [0.5ex] 
\hline 
breast-cancer & 99\% & 96\%   \\ 
credit & 99\%  & 99\%   \\
pasture & 99\%  & 98\%   \\
squash-unstored & 98\%  & 98\%   \\ 
squash-stored & 99\%  & 98\%   \\
white clover & 94\%  & 94\%   \\
eucalyptus & 99\%  & 98\%   \\
soybean & 99\%  & 97\%   \\
diabetes & 96\%  & 96\%   \\
glass & 99\%  & 99\%   \\
car & 99\%  & 99\%   \\
sonar & 99\%  & 99\%   \\
vehicle & 99\%  & 98\%   \\ 
\hline 
\end{tabular}}
\label{table:nonlin} 
\end{table}

\subsection{Performance Comparison with Pruned Neural Network Ensemble}

In a research by Lazarevic and Obradovic \cite{lazarevic2001effective} where clustering was also used to prune neural network ensembles, the researchers used diabetes and glass datasets, which we also used in our experiments. Table \ref{rivaltable1} depicts the accuracy of their entire and pruned ensembles with RF and our CLUB-DRF. For both datasets, our CLUB-DRF was superior to their pruned ensemble. Notably with the \emph{glass} dataset, CLUB-DRF has made a clear progression in predictive accuracy with a healthy 7.06\% increase over the pruned neural network. 

\begin{table}[lh]
\centering
\caption{Performance comparison between entire and pruned ensemble \cite{lazarevic2001effective} with our RF and CLUB-DRFs}
\label{rivaltable1}
\scalebox{0.9}{
\begin{tabular}{|c|c|c|c|c|}
\hline
Dataset &  Entire Ensemble &  Pruned Ensemble & RF & CLUB-DRF \\
\hline
Diabetes &   77.1\%   &                   77.9\%   &               81.26\%    &         81.42\%  \\ 
\hline
Glass    &   68.6\%           &           69.1\%     &             67.53\%      &      76.16\% \\
\hline
\end{tabular}}
\end{table} 

\subsection{Bias/Variance Analysis}
\label{bandv}
Bias and variance are measures used to estimate the accuracy of a classifier \cite{kohavi1995study}.  The bias measures the difference between the classifier's predicted class value and the true value of the class label being predicted. The variance, on the other hand, measures the variability of the classifier's prediction as a result of sensitivity due to fluctuations in the training set. If the prediction is always the same regardless of the training set, it equals zero. However, as the prediction becomes more sensitive to the training set, the variance tends to increase. For a classifier to be accurate,  it should maintain a low bias and variance. 

There is a trade-off between a classifier's ability to minimize bias and variance. Understanding these two types of measures can help us diagnose classifier results and avoid the mistake of over- or under-fitting. Breiman et al. \cite{leo1984classification} provided an analysis of complexity and induction in terms of a trade-off between bias and variance. 
In this section, we will show that CLUB-DRF can have a bias and variance comparable to and even better than RF. Starting with bias, the first column in Table \ref{biasPrunLevel} shows the pruning level of CLUB-DRF that performed the best relative to RF, and the second column shows the pruning level of the smallest CLUB-DRF that outperformed RF. As demonstrated in the table, CLUB-DRF has outperformed RF on all datasets. On the other hand, Table \ref{variancePrunLevel} shows similar results but variance-wise. Once again, CLUB-DRF has outperformed RF on all datasets. Although looking at bias in isolation of variance (and vice versa) provides only half of the picture, our aim is to demonstrate that with a pruned ensemble, both bias and/or variance can be enhanced. We attribute this to the high diversity our ensemble exhibits. 

\begin{table}[ht]
\caption{Pruning Level for CLUB-DRF Bias} 
\label{biasPrunLevel}
\scalebox{0.9}{
\centering 
\begin{tabular}{l c c c} 
\hline\hline 
Dataset & Best Performer & Smallest CLUB-DRF Outperforming RF \\ [0.5ex] 
\hline 
breast-cancer & 97\% & 97\%   \\ 
audit & 99\% & 99\%   \\
credit & 96\%  & 99\%   \\
pasture & 92\%  & 96\%   \\
squash-unstored & 99\%  & 99\%   \\
squash-stored & 93\%  & 97\%   \\
white clover & 99\%  & 99\%   \\
eucalyptus & 98\%  & 98\%   \\
soybean & 97\%  & 99\%   \\
diabetes & 99\%  & 99\%   \\
glass & 98\%  & 98\%   \\
car & 99\%  & 99\%   \\
sonar & 98\%  & 99\%   \\
vehicle & 99\%  & 99\%   \\
vote & 99\%  & 99\%   \\ 
\hline 
\end{tabular}}
\label{table:nonlin} 
\end{table}

\begin{table}[ht]
\caption{Pruning Level for CLUB-DRF Variance} 
\label{variancePrunLevel}
\scalebox{0.9}{
\centering 
\begin{tabular}{l c c c} 
\hline\hline 
Dataset & Best Performer & Smallest CLUB-DRF Outperforming RF \\ [0.5ex] 
\hline 
breast-cancer & 98\% & 98\%   \\ 
audit & 98\% & 98\%   \\
credit & 99\%  & 99\%   \\
pasture & 98\%  & 99\%   \\
squash-unstored & 98\%  & 99\%   \\
squash-stored & 95\%  & 98\%   \\
white clover & 95\%  & 98\%   \\
eucalyptus & 95\%  & 98\%   \\
soybean & 97\%  & 99\%   \\
diabetes & 97\%  & 98\%   \\
glass & 93\%  & 98\%   \\
car & 95\%  & 98\%   \\
sonar & 98\%  & 98\%   \\
vehicle & 99\%  & 99\%   \\
vote & 99\%  & 99\%   \\ 
\hline 
\end{tabular}}
\label{table:nonlin} 
\end{table}

\begin{figure}[H]
\scalebox{0.60}{
\centering
\includegraphics[width=6in,height=4in]{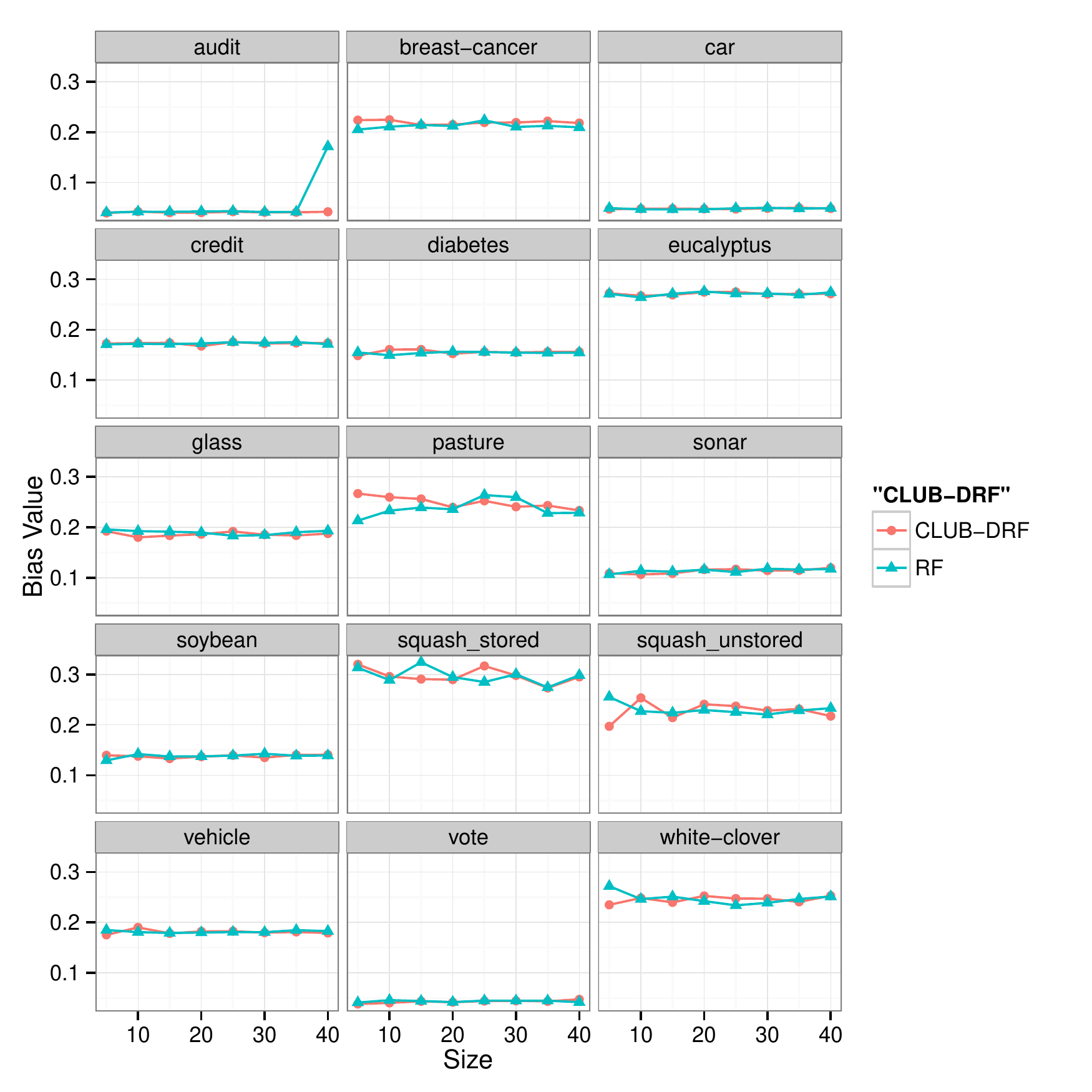}}
\caption{Bias Comparison of CLUB-DRF and Random Forest}
\label{biasComp}
\end{figure}

\begin{figure}[H]
\scalebox{0.60}{
\centering
\includegraphics[width=6in,height=4in]{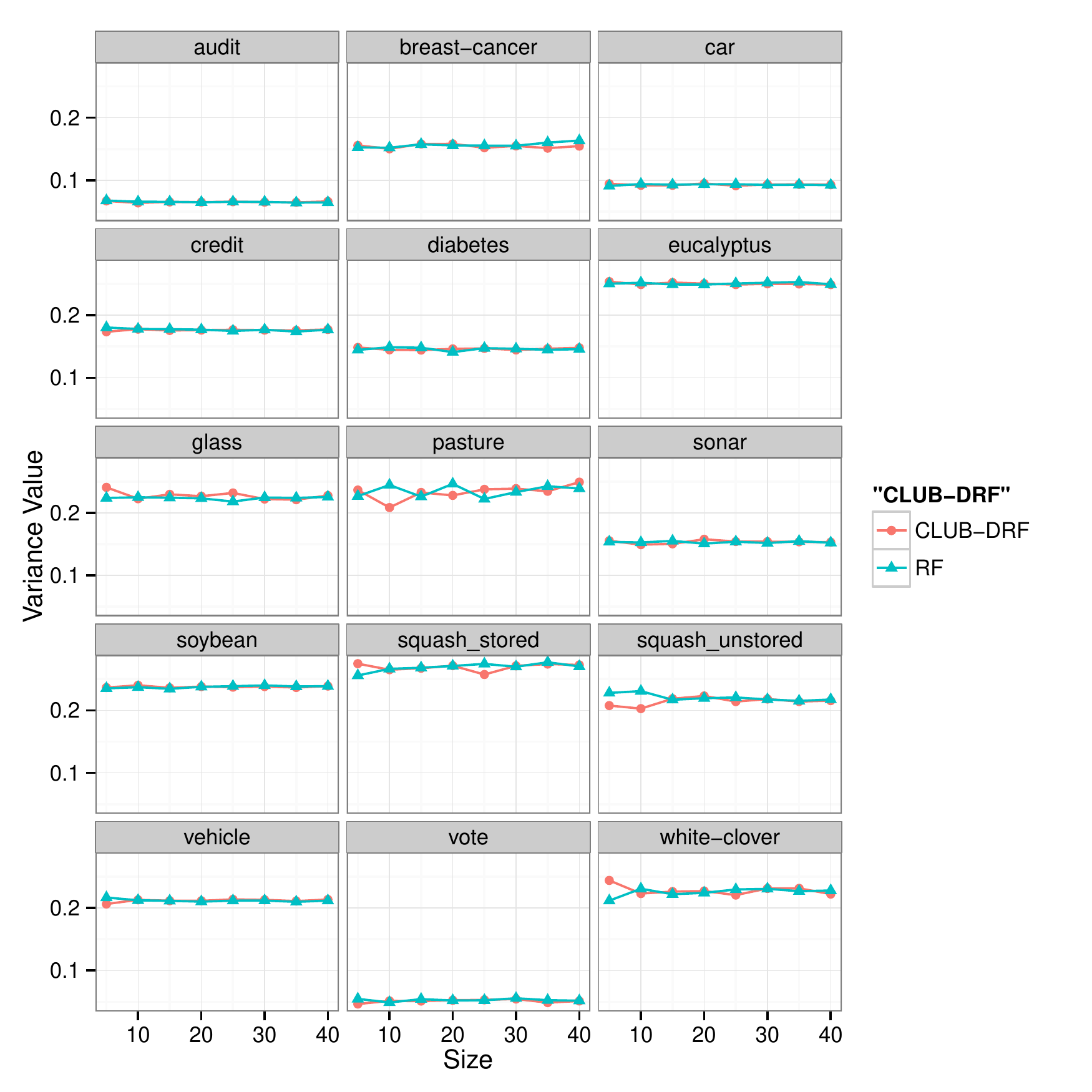}}
\caption{Variance Comparison of CLUB-DRF and Random Forest}
\label{varianceComp}
\end{figure}

We have also conducted experiments to compare the bias and variance between CLUB-DRFs and Random Forests of identical size. Figure \ref{biasComp} compares the bias and Figure  \ref{varianceComp} compares the variance. Both figures show that CLUB-DRF in most cases can have bias and variance equal to or better than Random Forest.

\section{Conclusion and Future Directions}
\label{conc}

Research conducted in this paper was based on how diversity in ensembles tends to yield better results \cite{kuncheva2003measures}  \cite{brown2005diversity} \cite{adeva2005accuracy} \cite{tang2006analysis}. We have used clustering to produce groups of similar trees and selected a representative tree from each group. These likely diverse trees are then used  to produce a pruned ensemble of the original one which we called CLUB-DRF.  As was demonstrated in the experiments section, CLUB-DRF, in most cases, performed at least as good as the original ensemble.

As a future research direction, we can consider using another method when selecting representative trees from the clusters. Instead of picking the best performing representative, a randomly selected representative from each cluster can be picked instead. It would be interesting to compare the results of both methods to determine the one that yields better results.  We can also try different sizes for the initial RF and also different cluster increments.

Another interesting research direction would be to use other clustering algorithms other than WEKA's own like DBSCAN \cite{ester1996density}, CLARANS \cite{ng2002clarans}, BIRCH \cite{zhang1996birch}, and/or CURE \cite{guha1998cure}.  Perhaps the way clusters are formed by each algorithm may have an impact on the performance of CLUB-DRF. This can happen when representatives selected from the clusters of one algorithm are more/less diverse than others selected from clusters produced by another algorithm.

\onecolumn
\footnotesize

\begin{center}

\begin{longtable}{|l||l||l||l||l||l||l||l||l||l|}

\caption[Performance Metrics of CLUB-DRF \& RF]{Performance Metrics of CLUB-DRF \& RF}
\label{bestrepoobtable} \\
\hline \multicolumn{1}{|c|}{\textbf{CLUB-DRF Size}} & \multicolumn{1}{c|}{\textbf{AVG}} & \multicolumn{1}{c|}{\textbf{MIN}} & \multicolumn{1}{c|}{\textbf{MAX}} & \multicolumn{1}{c|}{\textbf{SD}} & \multicolumn{1}{c|}{\textbf{Fmeasure}} & \multicolumn{1}{c|}{\textbf{AUC}} & \multicolumn{3}{|c|}{\textbf{AVG FMeasure AUC}} \\ \hline 
\endfirsthead

\multicolumn{3}{c}%
{{\bfseries \tablename\ \thetable{} -- continued from previous page}} \\
\hline \multicolumn{1}{|c|}{\textbf{CLUB-DRF Size}} & \multicolumn{1}{c|}{\textbf{AVG}} & \multicolumn{1}{c|}
{\textbf{MIN}} & \multicolumn{1}{|c|}{\textbf{MAX}} & \multicolumn{1}{|c|}{\textbf{SD}} & \multicolumn{1}{|c|}
{\textbf{Fmeasure}} & \multicolumn{1}{|c|}{\textbf{AUC}} & \multicolumn{3}{|c|}{\textbf{AVG FMeasure AUC}}
\\ \hline 
\endhead

\hline \multicolumn{3}{|r|}{{Continued on next page}} \\ \hline
\endfoot

\hline \hline
\endlastfoot
 \hline
\bf{breast-cancer} \\
 \hline
            5  &     \bf{71.24}  &    65.98    &  77.32    &   3.34   &   0.64    &  0.59  &  69.18 \hspace{2.5 mm} 0.63  \hspace{3 mm} 0.58 \\
  \hline
            10   &   \bf{72.16}   &   69.07    &  76.29     &  2.53    &  0.64   &   0.59 \\
 \hline
            15    &  \bf{70.21}    &  67.01    &  75.26     &  2.28   &   0.64    &  0.59 \\
 \hline
            20    &   \bf{72.58}    &  67.01    &  75.26     &  2.45   &   0.64   &   0.59 \\
 \hline
            25   &    \bf{69.59}    &  67.01    &  73.20     &  1.61   &   0.64   &   0.59 \\
 \hline
            30   &    \bf{71.65}   &   69.07    &  74.23    &   1.68   &   0.64   &   0.59 \\
 \hline
            35   &    \bf{69.48}   &   65.98    &  72.16    &   1.68   &   0.64   &   0.59 \\
 \hline
            40   &    \bf{71.44}   &   67.01    &  73.20    &   2.01    &  0.64    &  0.58 \\
 \hline
\bf{audit} \\
\hline
           5    &   95.86 &     93.22 &     97.05  &     1.19  &    0.92  &    0.88  &  96.53 \hspace{2.5 mm}   0.92  \hspace{3 mm}   0.88 \\
\hline
            10  &    96.18  &    94.99  &    96.76  &     0.49  &    0.91  &    0.87 \\
\hline
            15   &   96.06   &   95.43  &    96.76  &     0.50  &    0.91   &   0.88 \\
\hline
            20   &   96.47   &   95.58  &    97.05  &     0.41  &    0.91   &   0.87 \\
\hline
            25   &   96.30   &   95.87  &   96.61   &    0.31   &   0.91  &    0.87 \\
\hline
            30  &    96.42  &    96.17  &    96.61   &    0.15  &    0.91 &     0.87 \\
\hline
            35   &   96.39  &    95.72  &    96.61   &    0.25  &    0.91  &    0.87 \\
\hline
            40   &   96.42  &    96.17   &   96.76  &     0.18  &    0.91  &    0.87 \\
\hline
\bf{credit} \\
 \hline
  5   &     \bf{79.68}  &    72.06  &    87.94   &    5.06  &    0.68  &    0.62  &  77.47  \hspace{2.5 mm}  0.67  \hspace{3 mm}   0.61  \\
 \hline
            10   &    \bf{78.74}  &    75.00 &     82.65   &    2.38  &    0.68  &    0.62 \\
 \hline
            15    &   \bf{77.97}    &  75.59   &   80.29   &    1.84   &   0.68   &   0.62 \\
 \hline
            20    &  76.24   &   73.24    &  78.53     &  1.60   &   0.67     & 0.62 \\
 \hline
            25    &  77.35   &   75.00   &   80.29    &   1.53   &   0.67    &  0.61 \\
 \hline
            30  &    77.00   &   75.29   &   78.82     &  1.29    &  0.67   &   0.61 \\
 \hline
            35   &   76.94   &   75.29    &  78.53    &   1.02    &  0.67   &   0.61 \\
 \hline
            40   &   77.29    &  75.00    &  78.53    &   1.10   &   0.67   &   0.61 \\
 \hline 
\bf{pasture} \\
 \hline
      5   &     \bf{43.33}  &    16.67  &    66.67   &    15.72  &    0.45  &    0.55 &   41.67  \hspace{2.5 mm}  0.43  \hspace{3 mm}   0.57  \\
\hline
            10    &   \bf{55.83}  &    33.33   &   75.00  &     13.46  &    0.45  &    0.58 \\
\hline
            15   &    \bf{45.00}    &  16.67   &   66.67   &    15.90   &   0.43   &   0.56 \\
\hline
            20    &   \bf{49.17}  &    41.67  &    66.67   &    9.46    &  0.43   &   0.56 \\
\hline
            25  &     \bf{45.00}  &    16.67 &     58.33  &     13.02  &    0.44   &   0.55 \\
\hline
            30    &   \bf{45.83}   &   33.33  &    66.67  &     10.03 &     0.42  &    0.55 \\
\hline
            35   &    \bf{42.50}   &   25.00   &   58.33    &   10.17    &  0.44   &   0.57  \\
\hline
            40   &    \bf{45.00}   &   25.00  &    66.67  &     11.90 &     0.42 &     0.55 \\
 \hline
\bf{squash-unstored} \\
 \hline
     5     &   \bf{76.67}   &   50.00  &    100.00  &     12.25  &    0.63  &    0.72  &  72.50  \hspace{2.5 mm}  0.58  \hspace{3 mm}   0.70 \\
\hline
            10   &    \bf{80.00}   &   75.00   &   83.33   &    4.08   &   0.64   &   0.73 \\
\hline
            15   &   70.83   &   58.33   &   83.33     &  10.03   &   0.62   &   0.72 \\
\hline
            20   &    \bf{79.17}   &   66.67   &   91.67    &   7.68   &   0.60   &   0.72 \\
\hline
            25   &    \bf{75.83}   &   66.67   &   83.33    &   4.49    &  0.61    &  0.72 \\
\hline
            30   &    \underline{72.50}    &  50.00   &   83.33     &  12.94    &  0.59   &   0.70 \\
\hline
            35   &    \bf{75.00}    &  58.33   &   91.67    &   9.13    &  0.59    &  0.70 \\
\hline
            40   &   70.00   &   50.00   &   83.33    &   12.47    &  0.58    &  0.70 \\

 \hline

\hline
\bf{squash-stored} \\
\hline
          5   &     \bf{63.33}  &    50.00  &    91.67  &     13.02  &    0.57   &   0.64 &   55.83 \hspace{2.5 mm}  0.50  \hspace{3 mm}   0.58 \\
\hline 
           10   &    \bf{65.00}    &  50.00   &   83.33    &   9.72   &   0.55   &   0.62 \\
\hline
            15    &  54.17    &  41.67    &  66.67    &   8.54    &  0.51   &   0.59 \\
\hline
            20    &   \bf{65.00}    &  58.33    &  83.33    &   7.26    &  0.54   &   0.62 \\
\hline
            25    &  53.33   &   41.67    &  58.33    &   6.67    &  0.52   &   0.61 \\
\hline
            30    &  55.00   &   50.00   &   66.67    &   6.67   &   0.52   &   0.60 \\
\hline
            35    &  52.50    &  41.67   &   66.67    &   6.51  &    0.50   &   0.59 \\
\hline
            40    &   \bf{56.67}   &   50.00   &   66.67    &   6.24   &   0.52   &   0.60 \\
 \hline
\bf{white clover} \\
\hline

 5      & 75.00  &    57.14 &     92.86  &     10.23  &    0.62   &   0.58   & 77.14  \hspace{2.5 mm}  0.60   \hspace{3 mm}  0.55 \\
\hline
            10    &  75.71   &   50.00 &     85.71  &     11.16  &    0.62   &   0.57 \\
\hline
            15  &    67.86    &  57.14   &   85.71     &  10.23  &    0.61  &    0.55 \\
\hline
            20    &  72.14  &    57.14   &   78.57  &     7.46  &    0.60  &    0.55 \\
\hline
            25    &  72.14    &  57.14   &   85.71   &    10.81    &  0.62   &   0.57 \\
\hline
            30   &   \bf{79.29}   &   57.14   &   92.86  &     9.82   &   0.62 &     0.57 \\
\hline
            35     & 73.57   &   64.29   &   85.71    &   6.43  &    0.62    &  0.57 \\
\hline
            40   &   73.57   &   64.29   &   78.57    &   4.57   &   0.62    &  0.56 \\
 \hline

\bf{eucalyptus} \\
 \hline
  5  &      \bf{55.46}   &   45.40    &  61.96   &    5.78  &    0.52  &    0.71  &  48.40  \hspace{2.5 mm}  0.51 \hspace{3 mm}   0.70 \\
\hline
            10   &    \bf{56.20}    &  46.63   &   63.80   &    5.74   &   0.52   &   0.71 \\
\hline
            15   &    \bf{52.52}  &    47.24  &    59.51  &     3.45    &  0.52   &   0.71 \\
\hline
            20  &     \bf{50.74}   &   45.40    &  52.15  &     2.05  &    0.51   &   0.70 \\
\hline
            25    &   \bf{52.15}  &    47.85  &    55.21   &    1.96  &    0.52   &   0.71 \\
\hline
            30    &   \bf{49.26}  &    45.40   &   53.99     &  2.65    &  0.51  &    0.70 \\
\hline
            35    &   \bf{51.23}  &    47.85   &   55.21     &  2.00   &   0.51   &   0.70 \\
\hline
            40  &     \bf{50.18}  &    45.40   &   53.99   &    2.66    &  0.51  &     0.70 \\
 \hline

\bf{soybean} \\
 \hline
           5   &     \bf{83.97}  &     78.15  &    92.72  &     4.10  &    0.76   &   0.89   & 82.32  \hspace{2.5 mm}  0.75 \hspace{3 mm}    0.89 \\
\hline
            10   &    \bf{82.52}  &    80.13  &    84.77    &   1.57  &    0.76  &    0.89 \\
\hline    
        15   &    \bf{84.11}   &   80.79    &  88.08     &  2.05  &     0.76   &   0.89 \\
\hline
            20 &     82.05  &    78.81   &   84.11  &     1.86 &     0.75  &    0.89 \\
\hline
            25 &     82.05  &    79.47 &     84.77   &    2.00  &    0.75   &   0.89 \\
\hline
            30   &   81.32   &   76.16  &    84.77    &   2.33    &  0.75  &    0.88 \\
\hline
            35   &   82.05  &    79.47   &   84.11   &    1.55   &   0.75  &    0.88 \\
\hline
            40  &    81.99  &    78.81   &   84.11    &   1.44   &   0.75  &    0.89 \\
 \hline

\bf{diabetes} \\
 \hline
 5   &    80.80 &     74.71  &    84.29  &     3.53  &    0.72  &    0.68 &   81.26  \hspace{2.5 mm}  0.71   \hspace{3 mm}  0.67 \\
\hline          
  10   &   81.15   &   74.71   &   84.29    &   3.56    &  0.71 &     0.68 \\
\hline
            15  &    79.85  &    77.39  &    83.14   &    1.96  &    0.71  &    0.67 \\
\hline
            20  &     \bf{81.42}   &   79.31  &    83.14  &     1.24  &    0.71 &     0.67 \\
\hline
            25  &    80.96   &   78.93    &  82.76   &    1.31   &   0.71  &    0.67 \\
\hline
            30   &   80.88    &  78.54  &    82.76   &    1.14   &   0.71   &   0.67 \\
\hline
            35   &   79.81   &   77.39   &   81.99    &   1.40    &  0.71    &  0.67 \\
\hline
            40   &    \bf{81.38}  &    80.08   &   83.14    &   0.94   &   0.71   &   0.67 \\
 \hline

\bf{glass} \\
 \hline
  5    &    \bf{76.16}   &   64.38  &    84.93   &    5.62   &   0.65 &     0.76  &  67.53  \hspace{2.5 mm}  0.63  \hspace{3 mm}   0.75 \\
\hline
            10   &    \bf{68.36}   &   61.64  &    75.34  &     4.52   &   0.64   &   0.76 \\
\hline
            15  &     \bf{71.37}   &   65.75   &   78.08   &    3.65   &   0.64   &   0.76 \\
\hline
            20  &     \bf{70.82}  &    67.12  &    73.97  &     2.30  &    0.64   &   0.76 \\
\hline
            25   &    \bf{68.22}  &    64.38   &   71.23    &   2.01    &  0.63   &   0.75 \\
\hline
            30   &    \bf{70.41}    &  67.12   &   76.71  &     2.75   &   0.64   &   0.76 \\
\hline
            35  &     \bf{69.32}  &    67.12  &    72.60    &   1.86   &   0.63  &    0.76 \\
\hline
            40   &    \bf{68.77}    &  63.01   &   71.23   &    2.51  &    0.64  &    0.76 \\
\hline
\bf{car} \\
 \hline
 5   &     \bf{64.17}  &    62.41  &    67.52   &    1.33 &     0.56   &   0.78 &   62.26  \hspace{2.5 mm}  0.56  \hspace{3 mm}   0.78 \\
\hline          
  10   &    \bf{63.01}  &     61.56    &  64.29    &   0.75   &   0.56   &   0.78 \\
\hline
            15  &     \bf{62.36}  &    60.71   &   64.29  &     1.12  &    0.56  &    0.78 \\
\hline
            20  &     \bf{62.35}    &  61.22  &    63.78   &   0.82  &    0.56  &    0.78 \\
\hline
            25    &   \bf{62.69}    &  60.88    &  63.95    &   0.85   &   0.56    &  0.78  \\
\hline
            30     & 62.18    &  61.05   &   63.10    &   0.82  &    0.56  &    0.78  \\
\hline
            35   &   61.96    &  60.88   &   63.61    &   0.72  &    0.56    &  0.78 \\
\hline
            40    &  61.99    &  61.05   &   62.59   &    0.54  &     0.55 &     0.78 \\
 \hline

\bf{sonar} \\
 \hline
 5   &     \bf{12.25} &     7.04  &    18.31  &     3.34   &   0.26    &  0.00   & 0.14  \hspace{2.5 mm}  0.29  \hspace{3 mm}   0.00 \\
\hline
            10   &    \bf{9.15}  &    0.00    &  16.90   &    5.20   &   0.28  &    0.00  \\
\hline
            15   &    \bf{6.34}   &   0.00   &   14.08    &   4.47  &    0.29  &    0.00  \\
\hline
            20   &    \bf{3.38}   &   0.00   &   8.45    &   2.76    &  0.29 &     0.00 \\
\hline
            25   &    \bf{3.10}    &  0.00    &  7.04   &    2.42  &    0.28  &    0.00 \\
\hline
            30   &    \bf{1.83}   &   0.00    &  4.23   &    1.27   &   0.28   &   0.00 \\
\hline
            35   &    \bf{3.38} &     0.00  &    4.23    &   1.29 &     0.28  &    0.00 \\
\hline
            40  &     \bf{3.38}  &    0.00   &   9.86     &  2.69   &   0.28  &    0.00 \\
 \hline

\bf{vehicle} \\
 \hline
 5    &    \bf{72.01}  &    67.36   &   81.25  &     4.76  &    0.66   &   0.77  &  69.90  \hspace{2.5 mm}  0.65  \hspace{3 mm}   0.77 \\
\hline          
  10   &    \bf{72.43}   &   64.58   &   81.25    &   5.22   &   0.66    &  0.77 \\
\hline
            15  &     \bf{71.25}   &   67.71   &   76.04   &    2.76  &    0.65    &  0.77  \\
\hline
            20   &    \bf{71.49}   &   69.44   &   73.96  &     1.28  &    0.65  &    0.77 \\
\hline
            25  &     \bf{70.94}   &   69.44   &   72.92   &    1.42   &   0.65  &    0.77 \\
\hline
            30  &     \bf{71.25}  &    68.75  &    75.00   &    2.03  &    0.66  &    0.77 \\
\hline
            35   &    \bf{70.10}    &  67.71   &   71.53    &   1.12  &    0.65 &     0.77 \\
\hline
            40  &     \bf{70.45}   &   68.75   &   72.57    &   1.07  &    0.65    &  0.77 \\
 \hline

\bf{vote} \\
 \hline 
5    &   94.53   &   91.89   &   97.30   &    1.37  &    0.91 &     0.94 &   95.95  \hspace{2.5 mm}  0.91 \hspace{3 mm}   0.94 \\
\hline
            10     &  95.74  &    95.27   &   96.62    &   0.43  &    0.91  &    0.94 \\
\hline          
  15    &  95.27  &    93.92  &     96.62   &    0.80  &    0.91  &    0.94 \\
\hline
            20 &     95.61  &    94.59   &   96.62    &   0.54  &    0.91   &   0.94 \\
\hline
            25  &    95.74   &   94.59   &   95.95  &     0.43  &    0.91   &   0.94 \\
\hline
            30 &     95.81  &    95.27   &   96.62   &    0.41   &   0.91    &  0.94  \\
\hline
            35  &    95.68  &    95.27  &    95.95   &    0.33  &    0.91   &   0.94 \\
\hline
            40  &    95.81  &    94.59   &   96.62   &    0.51   &   0.91  &    0.94 \\
 \hline
\end{longtable}
\end{center} 
\twocolumn

\ifCLASSOPTIONcaptionsoff
  \newpage
\fi

\bibliographystyle{IEEEtran}

\bibliography{IEEEabrv,prl}

\end{document}